\documentclass{article}

\usepackage[sort, numbers]{natbib}
\usepackage{adjustbox}
\setcitestyle{numbers, square}
\usepackage[font=small,skip=3pt]{caption}
\usepackage{float}
\usepackage{epsfig}
\usepackage{graphicx}
\usepackage{amsmath}
\usepackage{amssymb}
\usepackage{algorithm}
\usepackage{algpseudocode}
\usepackage{ctable}
\usepackage{multirow}
\usepackage{epstopdf}
\usepackage{color, soul}
\usepackage{colortbl}
\usepackage{tabularx}
\usepackage{booktabs}
\usepackage{pifont}
\usepackage{subcaption}

\usepackage{listings}
\newcommand{\vect}[1]{\boldsymbol{#1}}

\definecolor{tabrow}{rgb}{0.88,0.88,0.88}
\definecolor{Gray}{gray}{0.8}

\usepackage[final]{nips_2017}

\usepackage[utf8]{inputenc} 
\usepackage[T1]{fontenc}    
\usepackage{hyperref}       
\usepackage{url}            
\usepackage{booktabs}       
\usepackage{amsfonts}       
\usepackage{nicefrac}       
\usepackage{microtype}      

\title{Personalized Gaussian Processes for Future Prediction of Alzheimer's Disease Progression }

\author{
  Kelly Peterson$^1$,  Ognjen (Oggi) Rudovic$^1$, Ricardo Guerrero$^2$ and Rosalind W. Picard$^1$\\
  $^1$Massachusetts Institute of Technology\\
  $^2$Imperial College London\\
  {\tt\small \{kellypet,orudovic,rwpicard\}@mit.edu, reg09@imperial.ac.uk}
}

\begin{document}

\maketitle
\begin{abstract}
In this paper, we introduce the use of a personalized Gaussian Process model (pGP) to predict the key metrics of Alzheimer's Disease progression (MMSE, ADAS-Cog13, CDRSB and CS) based on each patient's previous visits. We start by learning a population-level model using multi-modal data from previously seen patients using the base Gaussian Process (GP) regression. Then, this model is adapted sequentially over time to a new patient using domain adaptive GPs to form the patient's pGP. We show that this new approach, together with an auto-regressive formulation, leads to significant improvements in forecasting future clinical status and cognitive scores for target patients when compared to modeling the population with traditional GPs. 
\end{abstract}

\section{Introduction}

Recent advances in machine learning have introduced novel opportunities for personalized healthcare, shifting from ``one-size-fits-all'' population modeling towards personalized models providing tailored health profiles and diagnoses \cite{Toschi2016, deyati2013,simmons2012}. Personalized models can provide deeper insights into each individual's condition with the potential to revolutionize treatment of myriad diseases, including diseases for which no effective treatment currently exists, such as Alzheimer’s Disease (AD) \cite{tadpole2017,mueller2005,jack2008,schelterns2003}. 
Due to the complex nature of AD, prediction of AD symptom onset -- a critical component in conducting effective clinical trials -- remains a challenge \cite{mangialasche2010,cummings2006,schelterns2003}. Of hundreds of clinical trials, costing billions of dollars, fewer than $1\%$ have proceeded to the regulatory approval stage and none have managed to prove a disease-modifying effect \citep{tadpole2017,cummings2006}. More successes depend on an improved ability to accurately identify patients at early ages of the disease where treatments are most likely to be effective. Thus, developing models with improved ability to automatically predict patients' future AD-related metrics indicating disease progression -- and to do so as early as possible, especially before the emergence of clinical symptoms -- is an important step towards this end.


A great deal of research into AD biomarkers during the previous decade has been based on data from the Alzheimer’s Disease Neuroimaging Initiative (ADNI)\cite{weiner2017}. The primary goal of ADNI has been to test whether serial brain-imaging scans, biological markers, and clinical and neuropsychological assessment, can be combined to measure the progression from cognitively normal (CN) to mild-cognitive impairment (MCI) and early AD. ADNI data are highly heterogeneous and multi-modal, and include imaging (MRI, PET), cognitive scores, CSF biomarkers, genetics, and demographics (e.g. age, gender, race) \citep{tadpole2017}. Although the heterogeneous nature of this dataset lends itself to building powerful, informative multi-modal models, the dataset itself is very sparse, with different combinations of features missing for dataset patients. Partial records are missing for roughly $80\%$ of patients \citep{campos2015}. 
Recently, the view on AD diagnosis has shifted towards a more dynamic process in which clinical and pathological markers evolve gradually before diagnostic criteria are met.
Given the wide variability in available data per patient, inherent per-person differences, and the slowly changing nature of the disease, accurate prediction of AD progression is a significant and difficult challenge.  

Most existing approaches focus on modeling patients based on their clinical status (CS), i.e. the clinical diagnosis categorizing them into one of the 3 main stages of AD (CN, MCI, AD). CS is often modeled based on commonly used cognitive measures, e.g., the mini mental state examination (MMSE)\citep{folstein1975}, the Washington University Clinical Dementia Rating Sum of Boxes score (CDRSB)\citep{hughes1982}, and the AD Assessment Scale-Cognitive subtest (ADAS-Cog13)\citep{rosen1984}. Modeling of these surrogate measures of disease progression has been explored in a number of works 
\citep{delor13,jedynak12,schmidt_richberg15_a,schmidt-richberg16,guerrero2016,ito14,donohue14,gavidia2017,zhu2016,Doyle2014}. However, the majority of these focus on modeling biomarkers at the population level; for instance, estimating typical trajectories of markers over the full course of the disease to estimate current disease progress and progression rate \cite{schmidt_richberg15_a,schmidt-richberg16}. Guerrero et al. \cite{guerrero2016} used mixed effects modeling to derive global and individual biomarker trajectories for a training population, which was later used to instantiate subject-specific models for unseen patients. Some of the modeling techniques \cite{guerrero2016,schmidt_richberg15_a,schmidt-richberg16} require cohorts with known disease onset and are thus prone to bias due to the uncertainty of the conversion time.
While \citet{schmidt_richberg15_a,schmidt-richberg16} make no parametric assumption on the model shape, their approach relies on the population-only model; while  Guerrero et al. \cite{guerrero2016} estimate {\it parametrized} individualized models, based on the subpopulations similarities, to derive patient-tailored predictive models. While the latter approach is personalized, its parametric models are limited in their modeling power as they may not be able to fully represent the underlying multi-modal distribution of the highly heterogeneous patient data. 

\begin{figure}[t]
\centering
\includegraphics[scale=.35]{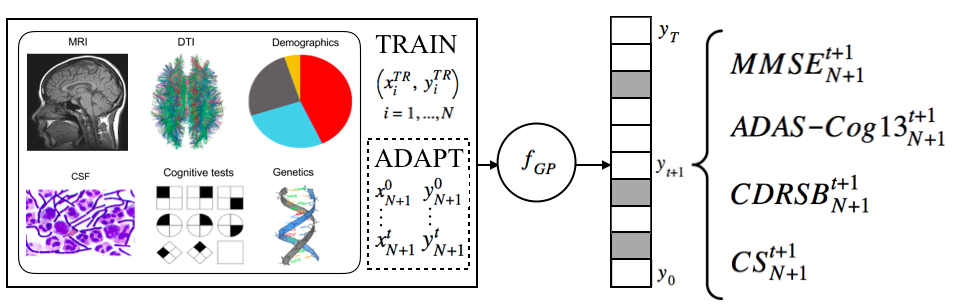}
\caption{{\small {\bf Personalized GPs.} The population model is first trained using all past visits data of $N$ patients ($x^{TR},y^{TR}$), where the time difference between two visits is 6 months. The model personalization to the target patient $(N+1)$ is then achieved by sequentially adapting the model predictions of the future metrics $y_{t+1}$ (using the posterior distribution of GPs - $f_{GP}$), informed by the visits data up to time step $t$. The shaded fields in the output represent the time points for which no visit data is available for a given patient.}}
\label{overview}
\end{figure}

Gaussian Process (GP) models have been shown to provide powerful probabilistic predictions for clinical analyses by using a non-parametric regression approach \citep{ziegler2014,lorenzi2017,lorenzi2015a,hyun2016}. For example,~\citet{ziegler2014} proposed a GP-based generative model allowing individualized predictions in patients at risk of developing dementia. 
Similarly, \citet{lorenzi2015a,hyun2016} used GPs for spatio-temporal modeling to delineate the developmental trajectories of brain structure and function in AD patients. While these methods exploit individual data to tune target models to specific patients, they do not provide a principled way of adapting the population model -- informed by the data of all training subjects -- to target new patients over time using data from their most recent clinical visits.



In this work, we introduce a personalized Gaussian Process model (pGP) to predict the key metrics of the AD progression (MMSE, ADAS-Cog13, CDRSB and CS) based on each patient's previous \textit{visits}. Here, a patient's \textit{visit} refers to the data collected at a single time point sample during the ADNI study (see Appendix D.3.). Specifically, in the pGP model, we start by learning a population model using multi-modal data of previously seen patients using the base GP regression~\cite{rasmussen2006gaussian}. Then, this model is adapted sequentially over time to a new patient using the notion of domain adaptive GPs~\cite{liu2015bayesian,eleftheriadis2017gaussian}. The key to our approach is a novel adaptation strategy for personalizing the GP population model (see Fig.\ref{overview}). We show this new approach leads to significant improvements in the prediction performance of the future clinical status and cognitive scores for target patients when compared to the population model.  

\section{Methods}
{\bf Notation.} We consider a supervised setting, where $\vect{X}^{(s)} = \{\vect{x}^{(s)}_{n_s}\}_{n_s=1}^{N_s}$ represents the multi-modal input features of $N_s$ patients from the ADNI dataset used to train the population model. The outcome scores for single patient visits are stored as $\vect{y}^{(s)}_{n_s}=\{\text{MMSE}^{(s)}_{n_s}\in\text{(0-30)}, \text{ADAS-Cog13}^{(s)}_{n_s}\in\text{(0-85)}, \text{CDRSB}^{(s)}_{n_s}\in \text{(0-18)},\text{CS}^{(s)}_{n_s}\in\{\text{CN=0,MCI=1,AD=2}\}\}$, and $\vect{Y}^{(s)}=\{\vect{y}^{(s)}_{n_s}\}_{n_s=1}^{N_s}$. Furthermore, each patient is represented by data pairs: $\{ \vect{x}^{(s)}_{n_s},\vect{y}^{(s)}_{n_s}\}$, where $\vect{x}^{(s)}_{n_s}=\{x_1,..,x_{t}\}$ contains the input features up to visit $t$ and $\vect{y}^{(s)}_{n_s}=\{y_2,..,y_{t+1}\}$ the corresponding target labels for the future visit $t+1$. The total number of visits is denoted by $T$. For notational convenience, we drop dependence on $n_s$, and use $\{x_{t}^{(s)},y_{t+1}^{(s)}\}_{t=1:T-1}$ as data pairs for learning the (population) prediction model. Since some patients missed certain visits and not all biomarkers were recorded at every visit, we fill in their missing values using their nearest available past visit, i.e., no data of future visits is used.
 
 {\bf Population-level GP.} We first build the population-level model using data of training subjects $\{\vect{X}^{(s)},\vect{Y}^{(s)}\}$. We formulate the predictions as a regression problem, where the goal is to predict patient's future outputs $y_{t+1}^{(s)}$ from data of previous visits, $x_{t}^{(s)}$ and $y_{t}^{(s)}$. To this end, we use the auto-regressive GP~\cite{candela2003propagation}, which we denote as GP(AR). This model tries to learn the following mapping function:
\begin{equation}
 y_{t+1}^{(s)} = f^{(s)}(x_{t}^{(s)};y_{t}^{(s)}) + \epsilon_{t}^{(s)},
\end{equation}
where $\epsilon_t^{(s)}\sim \mathcal{N}(0, \sigma^2_{s})$ is i.i.d. additive Gaussian noise. Following the framework of GPs~\cite{rasmussen2006gaussian}, we place a prior on the functions $f^{(s)}$. This gives rise to the joint prior $p(\vect{Y}^{(s)}_{2:T}|\{\vect{X}^{(s)},\vect{Y}^{(s)}\}_{1:T-1}) = \mathcal{N}(\vect{Y}^{(s)}_{2:T}|\vect{0}, \vect{K}^{(s)}_{1:T-1})$, where the elements of $\vect{K}^{(s)}_{1:T-1} = k^{(s)}(\{\vect{X}^{(s)},\vect{Y}^{(s)}\}_{1:T-1},\{\vect{X}^{(s)},\vect{Y}^{(s)}\}_{1:T-1})$ are computed using radial basis function (RBF) -isotropic kernel. 
The kernel parameters $\vect{\theta}$ were chosen to minimize the negative log-marginal likelihood: $-\log p(\vect{Y}^{(s)}_{2:T}|\{\vect{X}^{(s)},\vect{Y}^{(s)}\}_{1:T-1}, \vect{\theta})$. Given data of a {\it new} patient data from the visit at time $t$, $\vect{x}_\ast^{(s)}=\{x_{t}^{(s)},y_{t}^{(s)}\}$, the population GP predictive distribution provides the mean and variance estimate of the future metrics $y_{t+1}^{(s)}$ as:
\begin{align}
\small
\label{post_mu}\vect{\mu}_{t+1}^{(s)}(\vect{x}_\ast^{(s)}) &= {\vect{k}_\ast^{(s)}}^T (\vect{K}^{(s)} + \sigma_s^2\vect{I})^{-1}\vect{Y}^{(s)} \\ 
\label{post_s}\vect{V}_{t+1}^{(s)}(\vect{x}_\ast^{(s)}) &= k_{\ast\ast}^{(s)} - 
{\vect{k}_\ast^{(s)}}^T (\vect{K}^{(s)} + \sigma_s^2\vect{I})^{-1}
\vect{k}_\ast^{(s)},
\end{align}
where {\small $\vect{k}_\ast^{(s)} = k^{(s)}(\vect{X}^{(s)}, \vect{x}_\ast^{(s)})$} and {\small $k_{\ast\ast}^{(s)} = k^{(s)}(\vect{x}_\ast^{(s)}, \vect{x}_\ast^{(s)})$}, and the distribution mean is used as a point estimate of target outputs. For convenience, we denote $\vect{\mu}_{t+1}^{(s)} = \mu_{t+1}^{(s)}(\vect{x}_\ast^{(s)})$ and $V^{(v)}_{t+1} = V_{t+1}^{(s)}(\vect{x}_\ast^{(s)})$.

{\bf Personalized GP (pGP).}
\label{pgp}
We extend the approach of domain adaptive GPs (DA-GP)~\cite{liu2015bayesian,eleftheriadis2017gaussian} to personalize the population GP model to target patients. This is achieved by sequentially adapting the GP posterior for the test patient using the data of his/her past visits, to predict the future output metrics $y_{t+1}^{(p)}$. This is achieved by using the obtained posterior distribution of the source (population) data and data of the target patient up to visit $t$, to obtain a prior for the GP of the future data for the patient: $p(\vect{Y}^{(p)}_{t+1}|\{\vect{X}^{(p)},\vect{Y}^{(p)}\}_{1:t}, \mathcal{D}^{(s)}, \vect{\theta})$. Finally, this prior is used to correct the posterior distribution derived above to account for the previously seen data of the target patient. Formally, the conditional prior on the target patient data (given the source data) is given by applying Eqs.~(\ref{post_mu}--\ref{post_s}) on $\{\vect{X}^{(p)},\vect{Y}^{(p)}\}_{1:t}$ to obtain $\label{prior_mut}\vect{\mu}_{_{2:t+1}}^{(p|s)}$ and $\label{prior_st}\vect{V}_{2:t+1}^{(p|s)}$. Given this prior and a test input $\vect{x}_\ast^{(p)}=\{x_{t}^{(p)},y_{t}^{(p)}\}$, the correct form of the adapted posterior after observing the target patient data at visit $t$ is:\\
\scalebox{1}{\parbox{1\linewidth}{%
\begin{align}
\small
\label{post_muad}\mu_{t+1}^{(p)}(\vect{x}_\ast^{(p)}) &= \vect{\mu}_{t+1}^{(s)} +  {\vect{V}_{t+1}^{(p|s)}}^T (\vect{V}_{1:t}^{(p|s)} + \sigma_s^2\vect{I})^{-1}(\vect{Y}^{(p)}_{1:t} - \vect{\mu}_{1:t}^{(p|s)}) \\ 
\label{post_sad}V_{t+1}^{(p)}(\vect{x}_\ast^{(p)}) &= V_{t+1}^{(s)} - 
{\vect{V}_{t+1}^{(p|s)}}^T (\vect{V}_{1:t}^{(p|s)} + \sigma_s^2\vect{I})^{-1}
\vect{V}_{t+1}^{(p|s)},
\end{align}
}}\\
with {\small $\vect{V}_{t+1}^{(p|s)} = k^{(s)}(\vect{X}^{(p)}, \vect{x}_\ast^{(p)}) - 
{k^{(s)}(\vect{X}^{(s)}, \vect{X}^{(p)})}^T (\vect{K}^{(s)} + \sigma_s^2\vect{I})^{-1} k^{(s)}(\vect{X}^{(s)}, \vect{x}_\ast^{(p)})$}. Eqs.~(\ref{post_muad}--\ref{post_sad}) show that final prediction in the pGP is the combination of the original prediction based on the source data only, plus a correction term\footnote{For $t=1$, there is no correction term - the population model is used instead.}. The latter shifts the mean toward the distribution of the target patient and improves the model's confidence by reducing its predictive variance.

\section{Results}

Data used in this study were collected from the ADNI database \href{http://adni.loni.usc.edu/}{(adni.loni.usc.edu)}. We downloaded the standard dataset processed for the TADPOLE Challenge~\citep{tadpole2017}; this dataset represents 1,737 unique patients and was created from the \href{https://adni.bitbucket.io/index.html}{ADNIMERGE} spreadsheet, to which regional MRI (volumes, cortical thickness, surface area), PET (FDG, AV45, AV1451), DTI (regional means of standard indices) and cerebrospinal fluid (CSF) biomarkers were added. From this data, we constructed a multi-modal feature set consisting of six modalities: demographics (6 features), genetics (3 features), cognitive tests (9 features), CSF (3 features), MRI (366 features)(see Appendix D.1), and DTI (229 features). Due to sparseness, we excluded PET data entirely. For our experiments, we then selected a cohort of 100 patients with more than $10$ visits and missing no more than $82.5\%$ of the features (see Appendix D.2.1-2).

To evaluate performance, we ran a 10-fold patient-independent cross-validation. All the input features were z-normalized, and we then applied principal component analysis to reduce the effects of noise in the data, preserving $95\%$ of variance.
As performance metrics, we report the mean ($\pm$ SD) of the 10-folds, and in terms of commonly used metrics: mean absolute error (MAE) and intra-class correlation (ICC(3,1)\cite{shrout1979intraclass}. The latter ranges from 0--1, and is used to measure the (absolute) agreement between the model predictions and the ground-truth for target scores. For CS, we also compute the models accuracy (ACC) on three target labels (0--CN,1--MCI,2--AD). We report the results obtained using the population-level GPs, and their personalized versions (pGPs). Specifically, we compare the standard GPs and auto-regressive GPs -- GP(AR) -- the latter using also target scores as input features, but from previous visits of target patients. All GP models were trained to predict four target outputs, using the shared covariance function and the kernel parameters tuned on training data using GP toolbox~\cite{rasmussen2006gaussian}.

Table \ref{tab_res} shows the results. First, we see that GP(AR) significantly improves the prediction performance compared to standard GP (no past scores of the patients are used in the input). Thus, it appears that when learning the model's covariance function, including temporal changes in target scores significantly improves prediction results. The predictions by the personalized models (pGP/pGP(AR)) were obtained by first applying the population models (GP/GP(AR)), then correcting their predictions using the personalized adaption strategy introduced in Sec.\ref{pgp}. The result is that both personalized models outperform the population models by a large margin:  All improvements are statistically significant based on a paired t-test with equal variances and $p<0.01$. We also note significant improvements in accuracy: $24\%$ and $5\%$ on average. Fig.\ref{fig_cm}(See Appendix B) shows the confusion matrices for AR-models. We see pGP(AR) reduces the confusion between AD and MCI by $7.5\%$ across all patients, in addition to better predicting these two classes. Lastly, we show the improvements in the performance per patient, due to the model personalization, in Fig.\ref{fig:ind_per} (See Appendix B). 

\begin{table}
\begin{adjustbox}{width=1\textwidth}
\centering
\begin{tabular}{l|cccc|cccc|c}
  \hline
    \rowcolor{Gray}
{Models}        & & &\text{MAE} &  & & &\text{ICC(3,1)} &  &\text{ACC} \\
\hline
{}              & \begin{tabular}{@{}c@{}}\text{MMSE} \\\text{(0--30)}\end{tabular} & \begin{tabular}{@{}c@{}}\text{A-Cog13} \\\text{(0--85)}\end{tabular} &  \begin{tabular}{@{}c@{}}\text{CDRSB} \\\text{(0--18)}\end{tabular} & \begin{tabular}{@{}c@{}}\text{CS} \\\text{(0,1,2)}\end{tabular}
                & \begin{tabular}{@{}c@{}}\text{MMSE} \\\text{(0--30)}\end{tabular}
               & \begin{tabular}{@{}c@{}}\text{ADAS-Cog13} \\\text{(0--85)}\end{tabular} & 
               \begin{tabular}{@{}c@{}}\text{CDRSB} \\\text{(0--18)}\end{tabular} & 
               \begin{tabular}{@{}c@{}}\text{CS} \\\text{(0,1,2)}\end{tabular}
                & \begin{tabular}{@{}c@{}}\text{CS} \\\text{(0,1,2)}\end{tabular}\\
\hline
\toprule      
\textbf{GP}  &4.26$\pm$0.81	&6.10$\pm$1.29	&1.13$\pm$0.19 &0.43$\pm$0.08 
             &0.51$\pm$0.21 &0.71$\pm$0.06  &0.82$\pm$0.09 &0.66$\pm$0.11 &0.65$\pm$0.06\\       
     
\textbf{GP(AR)}  &2.70$\pm$0.52	&5.12 $\pm$1.16	&0.93$\pm$0.24 &0.23$\pm$0.04
             &0.70$\pm$0.12 &0.78$\pm$0.08 &0.87$\pm$0.07 &0.87$\pm$0.02 &0.87$\pm$0.03\\
\toprule             
                          
\textbf{pGP} &1.85$\pm$0.33	&4.43$\pm$0.83	&0.75$\pm$0.18 &0.19$\pm$0.04
             &0.81$\pm$0.14 &0.81$\pm$0.05  &0.90$\pm$0.05 &0.89$\pm$0.04 &0.89$\pm$0.04\\

\textbf{pGP(AR)}  &1.68$\pm$0.31	&4.30$\pm$0.83	&0.69$\pm$0.18 &0.14$\pm$0.03
                  &0.81$\pm$0.13    &0.83$\pm$0.08  &0.91$\pm$0.04 &0.92$\pm$0.02 &0.92$\pm$0.02\\
\bottomrule    
\end{tabular}
\end{adjustbox}
\caption{\small  Comparison of the population and personalized (p) models, and their auto-regressive (AR) variants.}
\label{tab_res}
\vspace*{-0.7cm}
\end{table}

\section{Conclusion}
\label{headings}
Timely diagnosis of AD is a challenging yet important problem whose solution could have significant impact on the success of diagnoses and disease-modifying treatments, especially in the earlier stages of the disease. This paper has proposed personalizing GP models (pGPs), and has tested two version of pGPs on a cohort of 100 patients from the ADNI dataset, showing significant improvements over population-level GP models. In the future, we plan to investigate how the sub-group similarities can be exploited to build patients' profiles, which can later be used to build local personalized models. We also plan to tackle the prediction of a longer time horizon, instead of one-visit-ahead, as done in this work. Extending this framework will help enable the prediction of changes in AD-related scores as early as possible; this capability is of great importance to both clinicians and those at risk of AD since it is critical to the early identification of at-risk patients, the construction of informative clinical trials, and the timely detection of AD progression.


\bibliography{egbib}

\vskip 5cm 
\appendix
{\bf \Large Appendix}

\section{Disease Progression Measured via Clinical Status Conversion}

\begin{table}[h]
\centering
\small
\begin{adjustbox}{width=1\textwidth}
\centering
\renewcommand{\arraystretch}{0.99}
\begin{tabular}{c|c|c|c|c|c|cc}
  \hline
    \rowcolor{Gray}
    {}  & {\small Conversion Type} & & & & &  &\\ \hline 
\toprule
{} & \text{\small CN $\rightarrow$ MCI$^{\ast}$} &  
\text{\small MCI$\rightarrow$ AD$^{\ast}$} 
 &  
\text{\small CN $\rightarrow$  AD} 
 &  
\text{\small MCI$\rightarrow$ CN}
 &  
\text{\small AD $\rightarrow$ MCI}
 &  
\text{\small AD $\rightarrow$ CN}
 \\ \hline 
{\text{\small $\#$ of Conversions}} & \small 14 & \small 27 & \small 1 & \small 3 & 
\small 1 & \small 0 \\ \hline
 
\bottomrule    
\end{tabular}
\end{adjustbox}
\vspace*{5mm}
\caption{\small Summary of the total number of observed changes in clinical status for the 100 patients used in the experiments conducted in this work. Change in clinical status of a patient is defined as the event when, given two sequential visits, the clinical status of the patient at those two visits differs.$^{\ast}$ The transition types denoted with asterisks (CN $\rightarrow$ MCI, MCI $\rightarrow$ AD) indicate key transition types associated with the progression of AD. }
\label{tab_res3}
\end{table}

Using clinical status (CS) as an indicator, we define patient conversion from one stage of Alzheimer’s Disease to another stage as the event when a patient’s CS at current visit at time $t$ is different than his/her CS at the previous visit $t-1$, $\forall t \in V$, where $\left\vert{V}\right\vert=n$, and $V$ is the set of all visits that a patient attends. The total number of possible visits that a patient may attend is denoted, $V_{possible}$, and $\left\vert{V_{possible}}\right\vert=22$. Note that it is possible that $\left\vert{V}\right\vert < 22$, since $\left\vert{V}\right\vert < \left\vert{V_{possible}}\right\vert$, i.e. the total number of visits a patient attends is not necessarily equal to the total number of possible visits, $V_{possible}$. 

Table 2 shows the frequency of patient conversions along the spectrum of AD progression, as measured by change in clinical status from one visit to the next. This information reflects the total number of conversions as observed across the 100 patients selected for use in our experiments. The transition types denoted with asterisks (CN $\rightarrow$ MCI, MCI $\rightarrow$ AD) indicate key transition types associated with the progression of AD.

\section{Observed Change in Cognitive Scores}

When considering the progression of AD, we are concerned with two patterns of clinical status conversion: 
\begin{enumerate}
  \item when a cognitively normal (CN) patient converts to MCI (CN $\rightarrow$ MCI)
  \item when an MCI patient converts to AD (MCI $\rightarrow$ AD). 
\end{enumerate}
These two scenarios represent the instances in which a patient’s clinical status progresses further along the spectrum of Alzheimer’s disease, with decreasing cognitive function.

In the figure below, we show the confusion matrices for predicting the future clinical status of the patients.

\begin{figure}[h]
\centering
\small
\setlength{\tabcolsep}{3pt}
\begin{tabular}{cc}
\includegraphics[scale=.15]{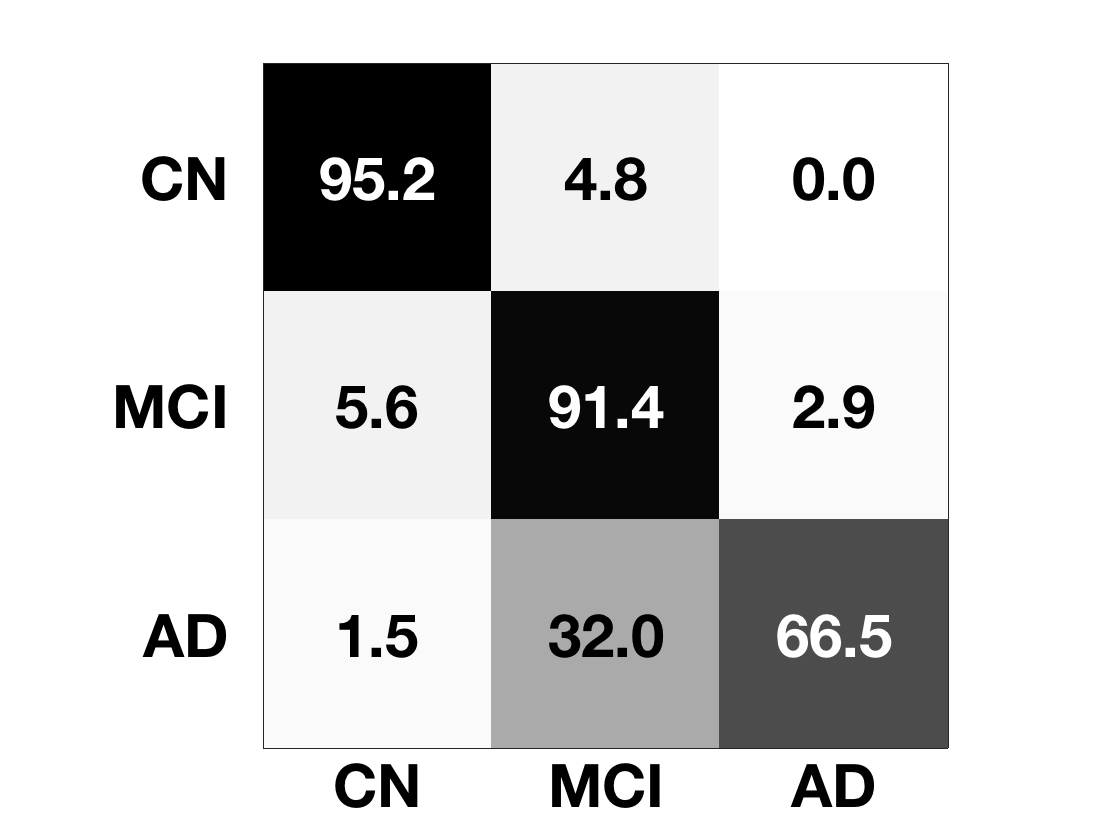} 
& \includegraphics[scale=.15]{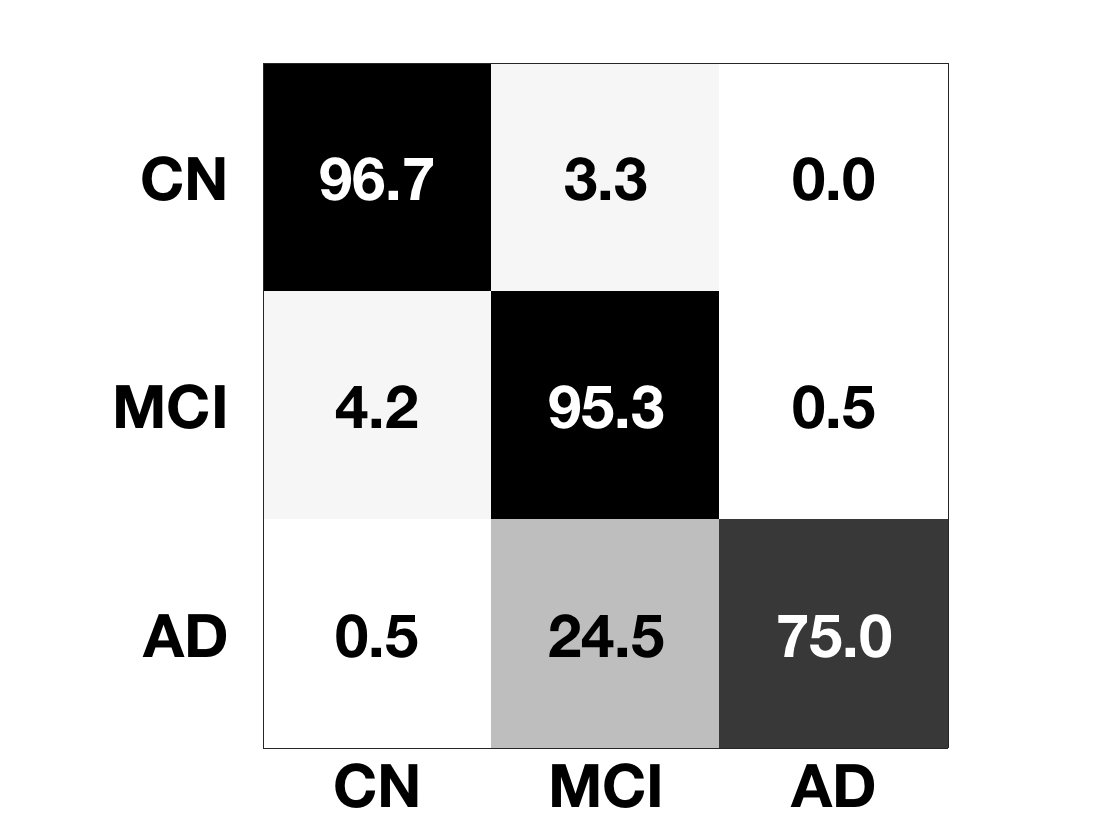}\\
GP(AR) & pGP(AR)
\end{tabular}
\caption{{\small Confusion matrices for predicting the {\it future} clinical status of the patients.}}
\label{fig_cm}
\vspace*{-.25cm}
\end{figure}

Analyzing the changes that occur when a patient undergoes conversion from one stage of AD to another is an important step toward understanding, characterizing, and profiling patients at each stage of the disease. In addition, it is necessary to compute the mean change in value (of metrics we wish to predict) that occurs when a patient undergoes a conversion and compare this mean delta with the MAE of our models; if the mean delta value is greater than the MAE obtained by our models, then we cannot assert that our models -- or any model -- accurately account for this mean change in value in cases when a patient changes clinical status. 

In this work, conversion is measured from one visit to the next. At the time $t$ of each conversion, we compute the change in cognitive score of the patient by calculating the absolute difference in cognitive score recorded at visit $t-1$ and the cognitive score recorded at visit $t$. For example, if a patient is diagnosed as MCI at the visit at time $t$, and the patient was diagnosed as CN at his/her previous visit at time $t-1$, then we compute the absolute difference in cognitive scores recorded at the visits at time $t-1$ and $t$. If a patient does not have cognitive scores recorded at the visit at $t-1$ and at $t$, then we do not calculate the absolute difference and do not consider this example. 

After computing all of the changes in value for each type of cognitive score (MMSE, CDRSB, ADAS-Cog13), we compute the mean and standard deviation of the change in cognitive score for each of the three cognitive tests. We define this value of the change in cognitive score as the delta ($\Delta$) of the cognitive score.

\begin{table}[h]
\begin{adjustbox}{width=1\textwidth}
\tiny
\centering
\begin{tabular}{l|cccc|cccc|c}
  \hline
    \rowcolor{Gray}
{Conversion Type} && \text{Mean Change in Cognitive Score} & & \\
\hline
{}              & \begin{tabular}{@{}c@{}}$\Delta$\text{ MMSE} \text{(0--30)}\end{tabular}  & \begin{tabular}{@{}c@{}}$\Delta$\text{ ADAS-Cog13} \text{(0--85)}\end{tabular} &  \begin{tabular}{@{}c@{}}$\Delta$\text{ CDRSB} \text{(0--18)}\end{tabular}\\
\hline
\toprule
\textbf{CN$\rightarrow$MCI}   &2.21$\pm$2.05 	&2.05$\pm$1.78 	&0.69$\pm$0.18 \\

\textbf{MCI$\rightarrow$AD}  &2.59$\pm$2.00	&1.63$\pm$1.31	&0.22$\pm$0.32\\
\bottomrule    
\end{tabular}
\end{adjustbox}
\caption{\small  Comparison of the mean change in cognitive score observed when patients convert to subsequent stages of AD. Mean and standard deviation were computed using data from the 100-patient cohort used in this work.}
\label{tab_res2}
\end{table}

In Table 3 shown above, we report the mean ($\pm$SD) of the delta ($\Delta$) value of each of 3 types of cognitive scores (ADAS-Cog13, CDRSB, MMSE) which our model tries to predict. The mean ($\pm$SD) is computed using data from all 100 patients. In the first row of Table 3, we provide the mean and standard deviation computed for all examples in which a patient converts from CN to MCI (CN $\rightarrow$ MCI). In the second row, we provide the mean and standard deviation computed for all examples in which a patient converts from MCI to AD (MCI $\rightarrow$ AD). Comparing these numbers with the achieved MAE by pGP(AR), we note that for MMSE and CDRSB the MAEs are $1.68$ and $0.69$, respectively. This suggests that the proposed model could detect subtle changes in these scores, which is one of the reasons for its significantly better detection of AD$\textbackslash$MCI states compared to the non-personalized model. \\


In ADNI, there is a baseline separation of about $2.2$ MMSE points between AD and MCI subjects, while there is a mean difference of $3.7$ MMSE points between MCI and CN subjects. When originally proposed, the MMSE~\cite{folstein1975mini} was shown to have a 24-hour test–re-test mean variation of $1.1$ points when the same tester did both examinations, and a mean variation of $1.3$ when different persons applied the test. Prediction errors made by the proposed method (MMSE MAE=$1.68$) is in fact close to the variability of the test itself.

\section{Individual Performance}
\begin{figure}
\centering
    \begin{subfigure}{.45\textwidth}
        \centering
        \includegraphics[width=\linewidth]{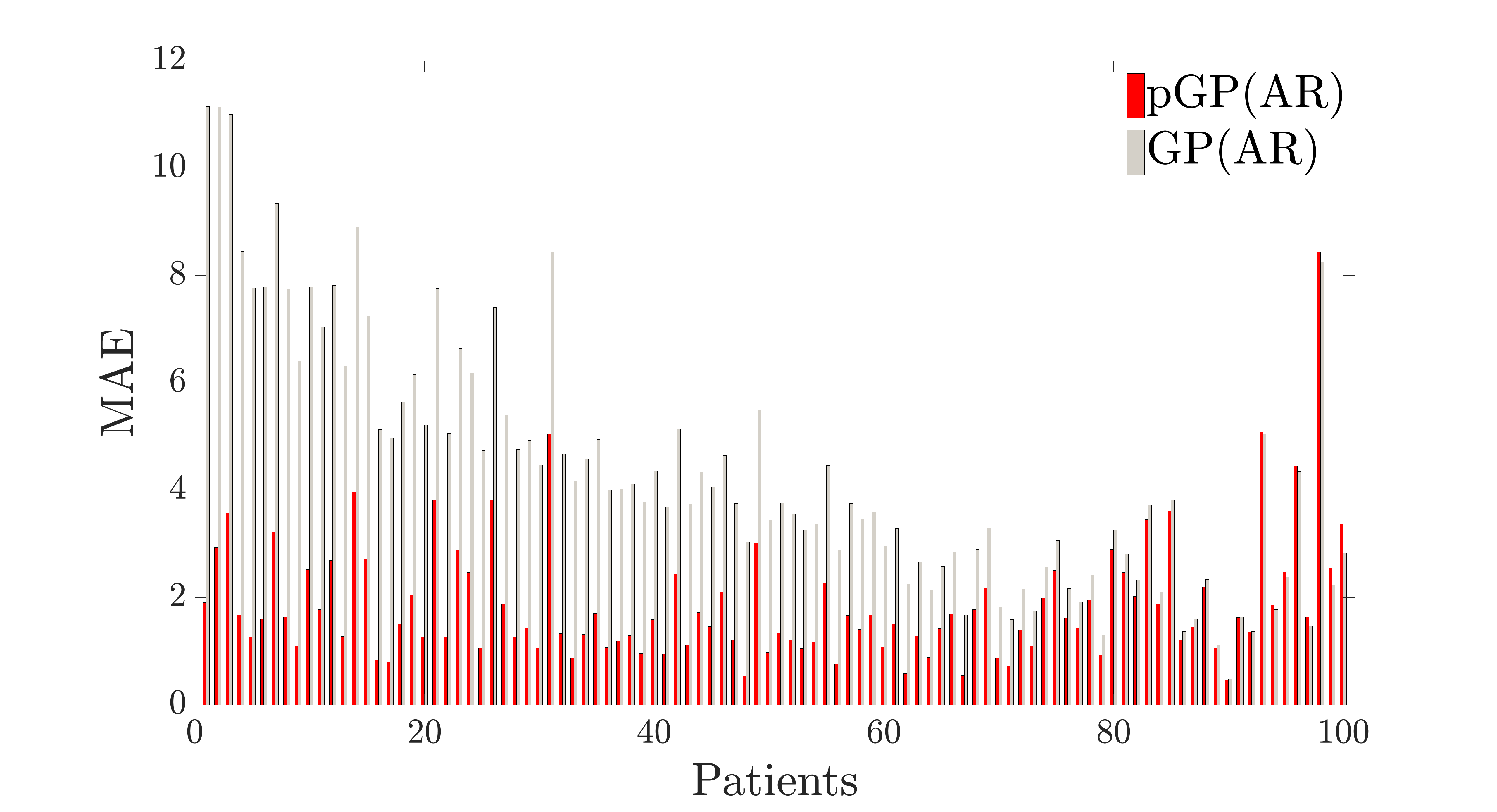}
        \caption{MMSE}\label{fig:fig_a}
    \end{subfigure} %
    \begin{subfigure}{.45\textwidth}
        \centering
        \includegraphics[width=\linewidth]{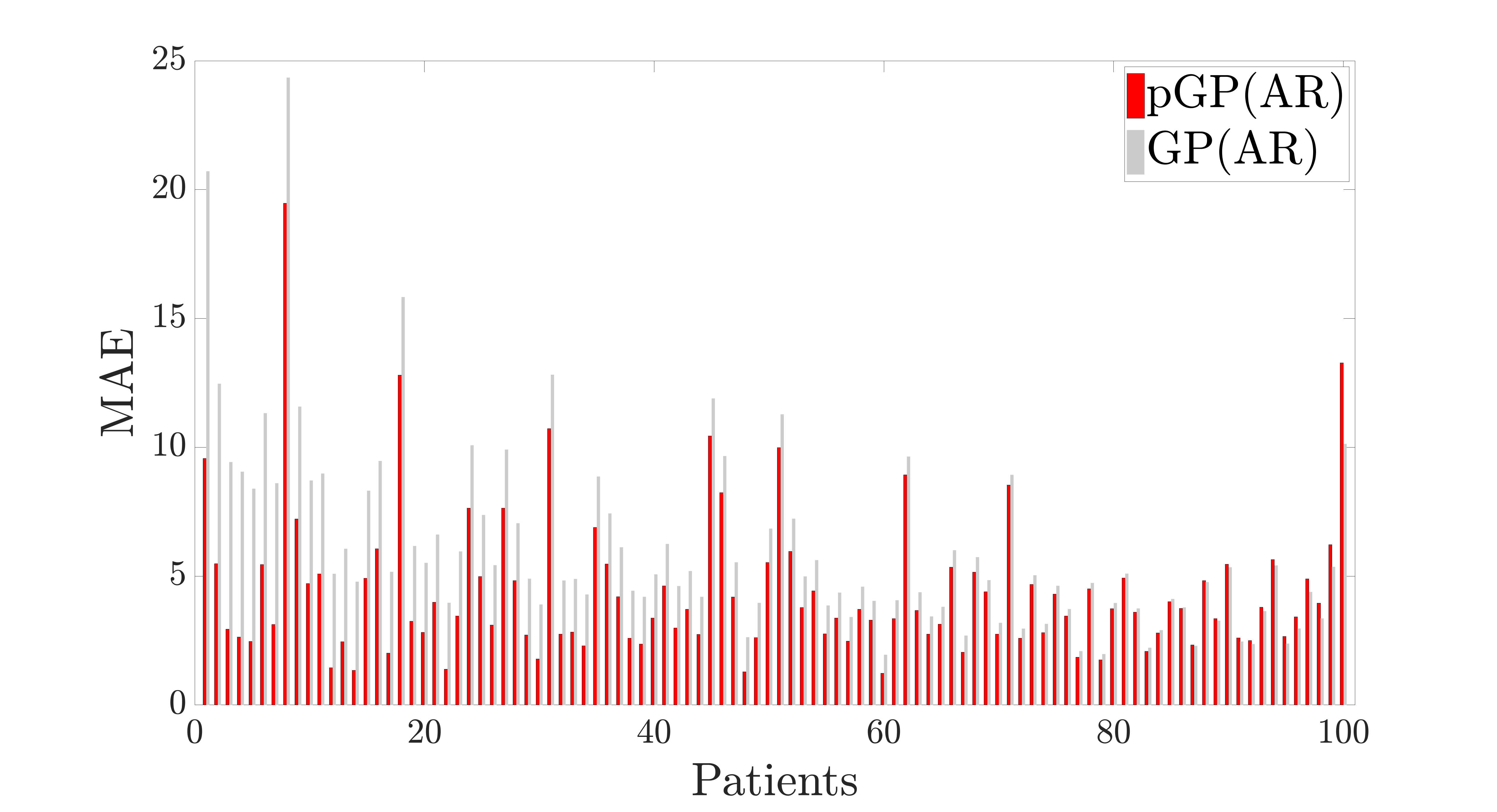}
        \caption{ADAS-Cog13}\label{fig:fig_b}
    \end{subfigure} %
    \begin{subfigure}{.45\textwidth}
        \centering
        \includegraphics[width=\linewidth]{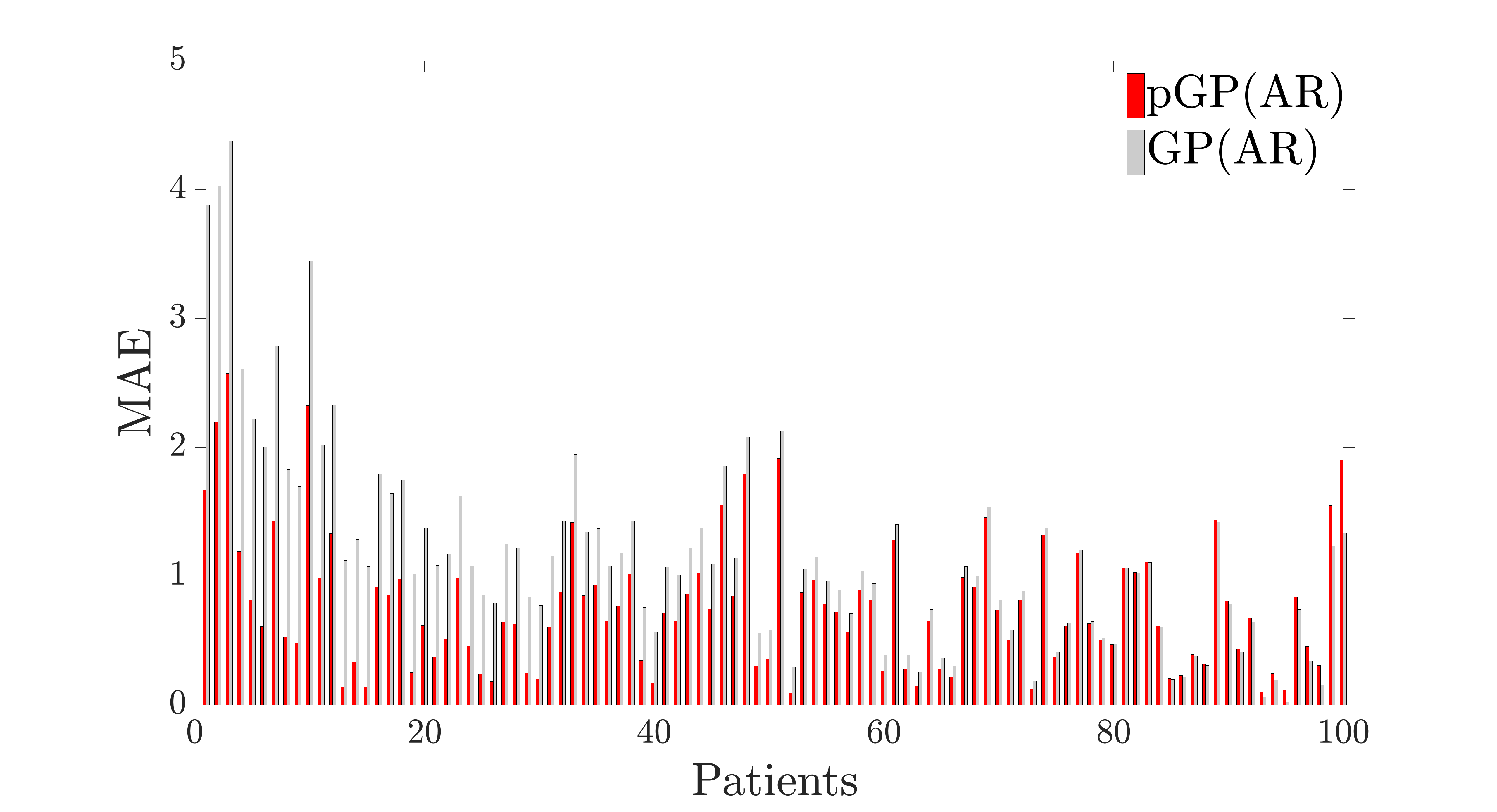}
        \caption{CDRSB}\label{fig:fig_c}
    \end{subfigure}
    \begin{subfigure}{.45\textwidth}
        \centering
        \includegraphics[width=\linewidth]{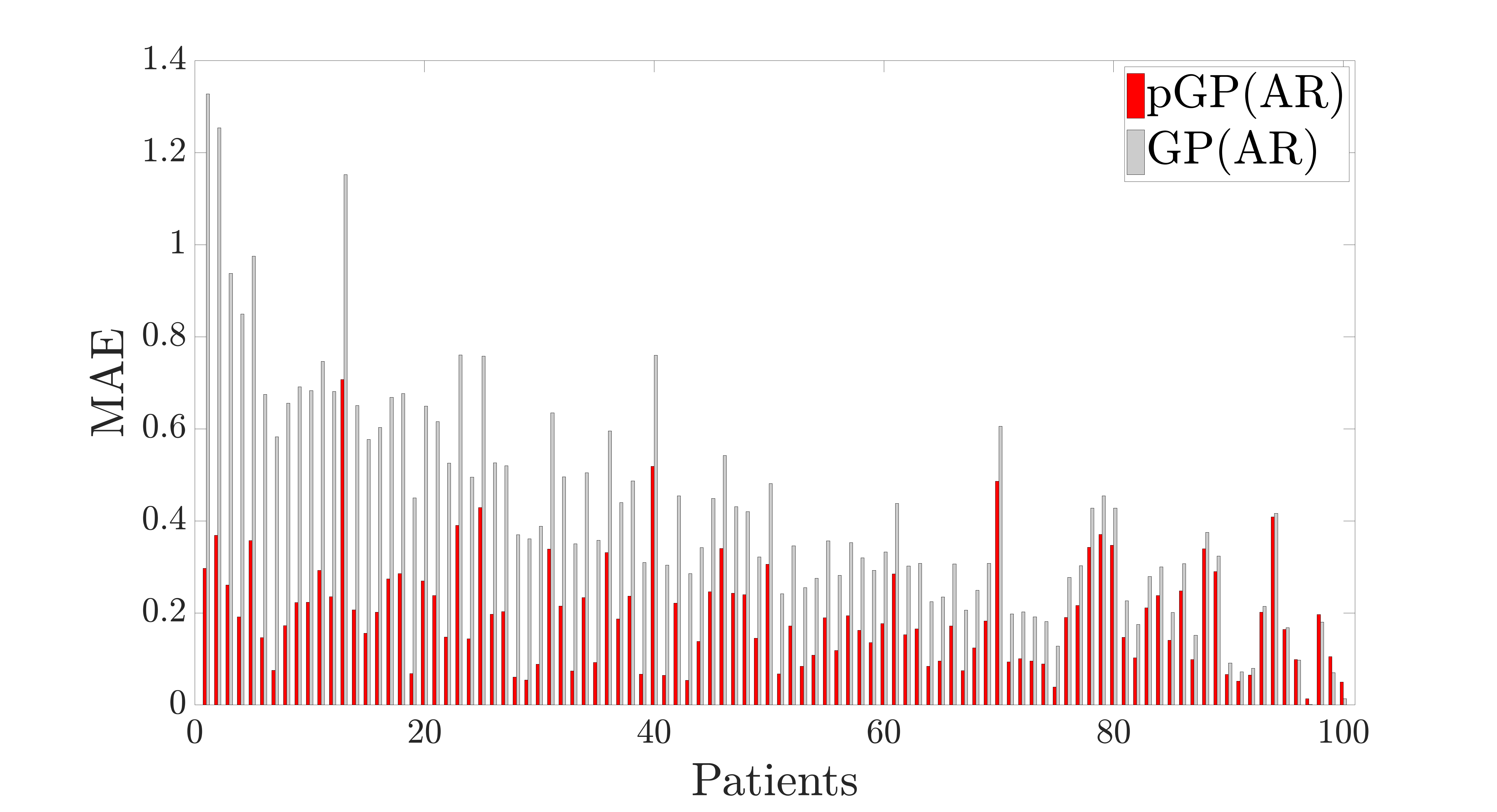}
        \caption{CS}\label{fig:fig_c}
    \end{subfigure}
\caption{The MAE errors for each of the target metrics computed per patient. The patients are sorted based on the relative improvements of personalized model - pGP(AR)- vs. non-personalized model - GP(AR), in a descending order. Overall, pGP(AR) outperforms GP(AR) on more than $80\%$ of patients across all four metrics.}
\label{fig:ind_per}
\end{figure}

Fig.\ref{fig:ind_per} show the individual performance for each patient, and for each output metric. As can be noted from the plots (a-d), the pGP(AR) achieves a lower MAE error than the GP(AR) on more than $80\%$ of all 100 patients across all four predicted metrics (MMSE, ADAS-Cog13, CDRSB, and CS). The relative performance improvements of the pGP(AR) compared to the GP(AR) are most readily evident in plot (a), when the models are used to predict MMSE. In plot (a), it is clear that that pGP(AR) achieves significant improvement in performance in comparison to the GP(AR); the MAE obtained by the pGP(AR) is much lower than that obtained by the GP(AR) for the majority of patients. The pGP(AR) also performs quite well in comparison to the GP(AR) when predicting CS (plot (d)). However, we note that the pGP(AR) and GP(AR) have more comparable performance when predicting ADAS-Cog13 and CDRSB (plots (c-d)). One possible explanation as to why the pGP(AR) is able to outperform the GP(AR) significantly when predicting MMSE, and yields more comparable performance results when predicting ADAS-Cog13 and CDRSB, may be due to the construction of cognitive tests and the inherent subjectivity of tests such as ADAS-Cog and its variants, as has often been noted in literature \cite{mohs1997}. Model-wise, we would like to emphasize that while both pGP(AR) and GP(AR) have seen the data of all visits from training patients, during inference the population-level model takes as input only the last visit. By contrast, the pGP(AR) actively updates and leverages its Bayesian framework to incorporate all the previous visit data of a test patient into its inference machinery. Evidently, this temporal personalization using the whole history of a test patient is critical for reducing the estimation error for the majority of the patients used in this work, as can be seen from Fig.\ref{fig:ind_per}.

\section{Data Preprocessing}
\subsection{Feature Selection}
Data used in this work was obtained from the ADNI database. Since its inception, ADNI has conducted three separate phases of study (ADNI-1, ADNI-GO, and ADNI-2) and recruited over 1,500 individuals, aged between 55-90 years, to participate in the study. As of November 2017, ADNI has begun a fourth phase of study called ADNI-3. Data collected from this phase of the study is not yet available in the ADNI database. 

The ADNI study aims to analyze biomarkers from cognitive tests, blood tests, tests of CSF, and MRI/DTI/PET imaging in order to understand and characterize the progression of Alzheimer’s disease. Thus, ADNI collects data samples for a myriad of these biomarkers and stores them in the ADNI database. 

In our work, we start by collecting a multi-modal dataset from the ADNI database, comprised of biomarkers across 7 different modalities:
\begin{enumerate}
  \item Cognitive tests 
  \item CSF
  \item MRI
  \item DTI
  \item PET
  \item Demographics
  \item Genetics
\end{enumerate}

Below, we provide a summary of each of the seven modalities, including information regarding the corresponding biomarkers available within each modality, and which biomarkers we selected to include in our feature set:\\

\textbf{\textit{1. Cognitive tests}}\\
Patients in the ADNI study were administered numerous cognitive tests (neuropsychological tests administered by a clinical expert) in order to diagnose AD progression. In the dataset collected from ADNI, we focus on data from 9 main cognitive tests (including subtypes). These include the following: 1) CDR Sum of Boxes (CDRSB), 2) ADAS-Cog11, 3) ADAS-Cog13, 4) MMSE, 5) RAVLT-immediate subtype, 6) RAVLT-learning subtype, 7) RAVLT-forgetting subtype, 8) RAVLT-percent forgetting subtype, and 9) FAQ. 

We use all nine cognitive tests as features. We also predict CDRSB, ADAS-Cog13, and MMSE.\\

\textbf{\textit{2. CSF}}\\
Cerebrospinal fluid (CSF) measurements are important for dementia research because they can be used to detect some of the earliest signs of AD. Analyzing the concentration of abnormal proteins such as amyloid-beta, tau, and phosphorylated tau provide strong evidence of progression of AD. ADNI provides three measures of CSF biomarkers: 1) amyloid-beta levels, 2) tau levels, and 3) phosphorylated tau levels. These measurements are not associated with a specific part of the brain \cite{tadpole2017}.

We use these 3 CSF-based biomarkers in our feature set. \\

\textbf{\textit{3. MRI}}\\
Magnetic resonance imaging (MRI) biomarkers can be used to measure damage to nerve cells. Specifically, MRI-based biomarkers can be used to quantify atrophy and structural brain integrity by measuring the volume of gray matter (GM) and white matter (WM) of the brain. \cite{tadpole2017}

TADPOLE datasets include three main types of structural MRI markers of atrophy: 1) ROI volumes 2) ROI cortical thicknesses 3) ROI surface areas, where an ROI (region of interest) is a 3D sub-region of the brain such as the inferior temporal lobe. These structural MRI markers were computed using an image analysis software called \href{https://surfer.nmr.mgh.harvard.edu/fswiki/LongitudinalProcessing}{Freesurfer} using two pipelines: cross-sectional (each subject visit is independent) or longitudinal (uses information from all the visits of a subject \cite{tadpole2017}.

From the provided MRI measurements, 18 measurements were null for all 1,737 patients We exclude these measurements from our feature set. In addition to the MRI volumetric features, we also used the manifold features from~\cite{guerrero2017group}. Using this set of MRI measurements, we obtain a total of 366 MRI-based features. \\

\textbf{\textit{4. DTI}}\\
Diffusion tensor imaging (DTI) measures the degeneration of white matter in the brain. In ADNI, DTI is a relatively recent imaging modality, and thus not all subjects have DTI scans. Three types of DTI ROI measurements were provided by ADNI: 1) mean diffusivity, 2) axial diffusivity, and 3) radial diffusivity. DTI measures are recorded for various voxels in the brain. \cite{tadpole2017}.

We obtain 229 DTI-based features using the DTI-based biomarkers provided for various brain ROIs of ADNI subjects. \\

\textbf{\textit {5. PET}}\\
Positron Emission Tomography (PET) imaging can be used to measure amyloid beta protein as well as to measure damage to nerve cells. In ADNI, three different types of PET measurements are provided: 1) AV45 PET ROI averages, 2) AV1451 PET ROI averages, and FG PET ROI averages. AV45 PET ROI measures amyloid-beta load in the brain, where amyloid-beta is a protein that misfolds, thus leading to AD. AV1451 PET ROI measures tau load in the brain; when tau load is abnormal, it causes damage to neurons, leading to AD. FDG-PET ROI measures cell metabolism, which is useful because AD-affected cells exhibit reduced metabolism \cite{tadpole2017}.

Due to sparseness, we do not include PET measurements in our feature set, as the majority of measurements were missing across all patients. 
\\

\textbf{\textit{6. Demographics}}\\
In addition to the above cognitive tests and quantitative biomarkers, ADNI provides demographic information for each patient. We use the following six types of demographic information as input features: 1) age, 2) gender, 3) ethnicity, 4) race, 5) years of education, and 6) marital status. \\

\textbf{\textit{7. Genetics}}\\
The alipoprotein E4 variant (APOE E4) is a gene that is the largest known risk factor for AD. Patients with APOE E4 have a high risk of developing AD. Each patient in ADNI underwent ApoE genotyping tests to determine if they have the APOE E4 gene present. In addition, each ApoE genotyping classifies patients into one of two genetypes (APGEN1 or APGEN2) based on their alleles \cite{tadpole2017}. These three pieces of genetic information are used as genetic-based features in our feature set.\\

After collecting data from ADNI, we performed additional data preprocessing steps in order to address missing values, select the features to use in our feature set, and to select the training set of patients.

As an initial step in handling missing or null values in the data from ADNI, we replaced all missing or null entries with a dummy value of $-99999999$, and convert all values to numeric format. After analyzing each modality and the percentage of missing values, we obtain a multi-modal feature set of $616$ features. In our model, we then address the remaining missing values by using previous values to fill in the missing ones. No future data that we aim to predict is used to fill in the missing values. 

We describe the steps to select the training set of patients in the following section. 

\subsection{Patient Subset Selection Criteria}
From our database of 1,737 patients, we selected a subset of patients to form the patient cohort used in our experiments. We selected patients to include in the cohort if they met the following two criteria: 
\begin{enumerate}
  \item  the patient attended more than $10$ separate visits (i.e. $11$ or more) during the 120-month period from which ADNI data was collected to our dataset
  \item  the patient is missing no more than $82.5\%$ of feature values
\end{enumerate}
Using these two criteria, we selected a subset of 100 patients. The application of these two criteria are described below, in Appendix D.2.1-2. The IDs of the 100 patients selected can be found at the end of this document (in Appendix D.4).

 \subsubsection{Visit History Criteria}
We used the frequency of each patient's visits as criteria to select a subset of patients in order to obtain a subset of patients for which sufficient data samples had been collected. As described in Appendix D.3, the majority of patients did not attend all $22$ possible visits during the ADNI study. Out of 1,737 patients, only 267 patients attended more than $10$ separate visits (i.e. at least approximately $50\%$ of all $22$ possible visits). We selected this subset of 267 patients with at least more than $10$ points in time at which data was collected using the hypothesis that the resulting dataset would not be too sparse such that imputation or filling in of missing values would be erroneous. Using this subset of 267 patients, we then selected our final cohort of 100 patients by computing the percentage of missing feature values available for each patient. This process is described in more detail in the following section, Appendix D.2.2.
 
 \subsubsection{Handling Missing Features}
From the subset of 267 patients selected as described in Appendix D.2.1, we then selected the subset of patients who were missing at most $82.5\%$ of feature data. In other words, a patient profile must contain at least $17.5\%$ of feature data, i.e. the profile must not be missing more than $82.5\%$ of feature data. For instance, if the elements in the table, containing the features of target patient for all the visits, had values $-99999999$, these were counted as the missing feature data for that patient. This high percentage of the missing data is not surprising since, for the majority of the patients, high dimensional modality data such MRI, PET and DTI, were not often present. Also, the latter two are relatively new imagining techniques, and only a few subjects in the TADPOLE dataset have undertaken these images \cite{tadpole2017}.


\subsection{Patient Visit Data Collection}

In the dataset used in this work, quantitative measurements for participants of the ADNI study were collected at clinical visits in six-month increments, on average. Time points were collected at various time points over the duration of 120 months, with samples collected for each patient from at most $22$ different time points. Each patient in the study has a data sample from at least one time point (i.e. clinical visit). The maximum number of samples collected per patient corresponds to the maximum number of clinical visits a patient attended over the course of the ADNI study. Thus, each patient has data collected from at most $22$ visits. Because patient participate in the ADNI study at varying time points and frequency, the maximum number of samples per patient (for the subset of patient considered here) is $19$. The average number of time point samples per patient (i.e. number of visits attended per patient) is $7.334$.

Patients who attended less than $5$ visits fall into the lower quartile ($25th$ quartile); patients who attended $9$ or more visits fall into the upper quartile ($75th$ percentile). The maximum possible number of visits a single patient may have attended is $22$. However, no patient attended all $22$ possible visits. Thus, we observe that the maximum number of visits a single patient in our dataset attended was $19$. Each patient attended a minimum of one visit. We note that $276$ or $15.88\%$ of all $1,737$ patients attended more than $10$ visits. Assuming a normal distribution with $\mu = 7.334$ and $\sigma = 4.033$, patients with more than $10$ visits fall into the $81st$ percentile. 

\begin{figure}[t]
\centering
\includegraphics[scale=.5]{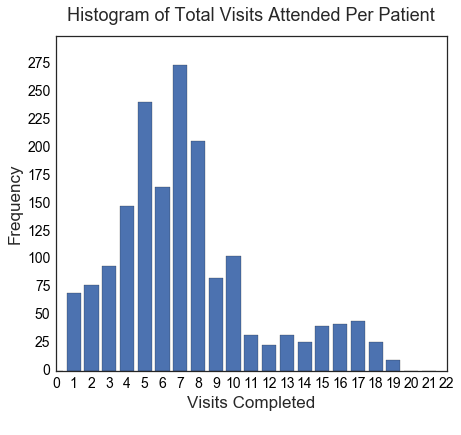}
\caption{{\small {\bf Histogram of Patient Visit Frequency.} Histogram showing the distribution of the total number of visits each of the 1,737 patients attended. The distribution of total patient visits has a mean, $\mu$, of $7.334$, and standard deviation, $\sigma$, of $4.033$. 
}}
\label{visit_hist}
\end{figure}



\subsection{Cohort Patient IDs}
Below is a list of the IDs of each of the 100 patients selected for use in the experiments performed in this work:\\\\
$21, 23, 31, 51, 61, 72, 89, 107, 108, 112, 120, 123, 126, 127, 130, 135, 142, 150, 169, 172,186, 200, \\
214, 256, 257, 259, 260, 269, 276, 295, 298, 301, 307, 331, 359, 361, 376, 378, 382, 384, 388, 413, 419, \\
441, 454, 539, 545, 546, 548, 610, 618, 626, 644, 649, 658, 668, 671, 679, 685, 698, 722, 729, 741, 752,\\
778, 800, 830, 835, 869, 887, 906, 919, 934, 952, 972, 984, 985, 994, 1045, 1072, 1078, 1097, 1098, \\
1122, 1123, 1155, 1186, 1187, 1232, 1246, 1261, 1268, 1269, 1300, 1318.$

\end{document}